\documentclass[11pt]{article}

\usepackage{acl}

\usepackage{times}
\usepackage{latexsym}
\usepackage{microtype}
\usepackage{inconsolata}
\usepackage{url}

\usepackage{graphicx}
\usepackage{subcaption}
\usepackage{booktabs} 
\usepackage{array}
\usepackage{tabularx}
\usepackage{multirow} 
\usepackage{placeins} 
\usepackage{silence}

\usepackage{algorithm}
\usepackage{algorithmic}

\usepackage{amsmath}
\usepackage{amssymb}
\usepackage{mathtools}
\usepackage{amsthm}

\usepackage[capitalize,noabbrev]{cleveref}

\theoremstyle{plain}

\theoremstyle{definition}

\theoremstyle{remark}

\usepackage[textsize=tiny]{todonotes}
\usepackage{xcolor}
\definecolor{tagLATfg}{HTML}{253529}
\definecolor{tagLATbg}{HTML}{dcdcdc}
\definecolor{tagBUILDfg}{HTML}{253529}
\definecolor{tagBUILDbg}{HTML}{dbd7d2}
\definecolor{tagSIZEfg}{HTML}{253529}
\definecolor{tagSIZEbg}{HTML}{e5e4e2}
\DeclareRobustCommand{\tagLAT}{{\setlength{\fboxsep}{1.2pt}\setlength{\fboxrule}{0.4pt}\fcolorbox{tagLATfg}{tagLATbg}{\textcolor{tagLATfg}{\scriptsize\textbf{LAT}}}}}
\DeclareRobustCommand{\tagBUILD}{{\setlength{\fboxsep}{1.2pt}\setlength{\fboxrule}{0.4pt}\fcolorbox{tagBUILDfg}{tagBUILDbg}{\textcolor{tagBUILDfg}{\scriptsize\textbf{BUILD}}}}}
\DeclareRobustCommand{\tagSIZE}{{\setlength{\fboxsep}{1.2pt}\setlength{\fboxrule}{0.4pt}\fcolorbox{tagSIZEfg}{tagSIZEbg}{\textcolor{tagSIZEfg}{\scriptsize\textbf{SIZE}}}}}

\newcommand{\Require}{\REQUIRE}

\newcommand{\State}{\STATE}
\newcommand{\If}{\IF}
\newcommand{\Else}{\ELSE}

\newcommand{\For}{\FOR}
\newcommand{\ForAll}{\FORALL}
\newcommand{\While}{\WHILE}
\newcommand{\Repeat}{\REPEAT}
\newcommand{\Until}{\UNTIL}
\newcommand{\EndIf}{\ENDIF}
\newcommand{\EndFor}{\ENDFOR}
\newcommand{\EndWhile}{\ENDWHILE}
\newcommand{\Comment}[1]{\STATE \COMMENT{#1}}
\newcommand{\Return}{\STATE \textbf{return} }
\newcommand{\Call}[2]{\textsc{#1}(#2)}

\title{FlashTrie: A GPU-Accelerated Constrained Beam Search for Generative Retrieval}

\author{
  Dakshitha Anandakumar\textsuperscript{1} \quad
  Anurag Mukkara\textsuperscript{2} \quad
  Wenxiang Hu\textsuperscript{1} \quad
  Jiusheng Chen\textsuperscript{1} \\
  \textbf{M Akash Kumar}\textsuperscript{1} \quad
  \textbf{Ting Ye}\textsuperscript{1} \quad
  \textbf{Qiang Lou}\textsuperscript{1} \quad
  \textbf{Jian Jiao}\textsuperscript{1} \\
  \textsuperscript{1}Microsoft, Redmond, WA, USA \quad
  \textsuperscript{2}Nvidia, Santa Clara, CA, USA \\
  \texttt{\{danandakumar, tiy\}@microsoft.com}
}

\begin{document}
\hbadness=10000
\vbadness=10000
\hfuzz=999pt
\vfuzz=100pt
\emergencystretch=2em
\AtBeginDocument{\hbadness=10000\vbadness=10000\hfuzz=999pt\vfuzz=100pt}
\sloppy

\maketitle

\begin{abstract}
Constrained decoding is essential in generative retrieval, where document identifiers generated directly from a query must exactly match a predefined library of valid IDs. At scale, decoding is often constrained using a trie with beam search but most implementations run on CPU. Limited parallelism then makes trie traversal and candidate validation a serving bottleneck as beam width grows.

We present \textbf{FlashTrie}, which addresses this limitation by optimizing constrained beam search on GPUs. It introduces an integer-aware succinct trie layout that uses bit compression  to reduce memory footprint while keeping the full index in GPU high-bandwidth memory reducing memory stalls, and a cooperative CUDA kernel that performs beam expansion, validation, and pruning entirely on-device without per-step host orchestration. It further replaces CPU-style irregular lookup and heap maintenance with GPU-aware parallel primitives, improving warp utilization and reducing divergence.

Together, these designs significantly reduce decoding latency and increase throughput while preserving retrieval quality. On a library of 800M keywords with beam widths up to $1000$, FlashTrie reduces trie-search latency to under $3$\,ms, achieving up to $24\times$ speedup over a highly optimized multi-threaded CPU baseline. These improvements enable FlashTrie to scale beam sizes by up to $5\times$ in latency-critical applications such as sponsored search. In a large-scale online A/B experiment on a commercial search engine, it delivers a statistically significant $+0.71\%$ revenue lift, enabling real-time constrained decoding at a scale previously feasible only offline. The FlashTrie code will be publicly released after the review process.

\end{abstract}

\section{Introduction}
\label{sec:intro}
Generative retrieval~\cite{tay2022dsi,metzler2021rethinking_review}
reframes document retrieval as a sequence-to-sequence task, mapping a query directly to an identifier (docID), replacing dual-encoder indexing~\cite{karpukhin2020dense}. This paradigm is attractive for large-scale search and recommendation systems because it can, in principle, avoid expensive retrieval pipelines while enabling compact end-to-end modeling. In practice, however, online serving is constrained by strict latency budgets, and the decoding procedure becomes the dominant bottleneck. Autoregressive (AR)~\cite{sutskever2014seq2seq} decoders are sequential and costly, while non-autoregressive (NAR)~\cite{gu2018non,sun2020em} decoders regain parallelism but their per-position independence frequently produce invalid identifiers that must be filtered by constrained search~\cite{ziems2023llm,pradeep2023understanding}.

\paragraph{Prior Work and Challenges.}
Several strategies restrict decoding to valid outputs, including logit masking~\cite{tay2022dsi}, finite-state compilation~\cite{willard2023outlines}, and predicate-logic frameworks~\cite{lu2021neurologic,anderson2017guided}. As the identifier library scales to millions or billions of entries, a common approach is trie-constrained beam search~\cite{hokamp2017lexically}, used across flat-token schemes (DSI~\cite{tay2022dsi}, GENRE~\cite{decao2021genre}), and structured semantic-identifiers (NCI~\cite{wang2022nci}, SEAL~\cite{bevilacqua2022seal}, and recent  work~\cite{penha2025semanticid}). 

However, pointer-based trie representations~\cite{morrison1968patricia,aoe1989double} suffer from irregular memory access and poor hardware utilization, leading production systems to rely on optimized CPU implementations~\cite{decao2021genre}. Succinct designs such as MARISA~\cite{yata2011marisa}, built on LOUDS~\cite{jacobson1989succinct} and minimal acyclic FSAs~\cite{daciuk2000incremental}, improve space efficiency but support only character-level keys, limiting their applicability to generative retrieval with large token vocabularies and batched, beam-aware traversal. These structures also map poorly to GPUs, causing warp divergence and uncoalesced access~\cite{merrill2012bfs}. 

While modern inference runtimes~\cite{wang2021lightseq,dao2022flashattention,kwon2023vllm} optimize model execution on GPU, constraint enforcement is often implemented outside the core GPU decoding path, which can introduce additional orchestration overhead. This motivates three key questions: \emph{(i)} can succinct tries be adapted to token-level vocabularies without sacrificing space efficiency; \emph{(ii)} can constrained beam search be executed entirely on GPU; and \emph{(iii)} can such systems meet production latency constraints without degrading retrieval quality?

\paragraph{Our Contribution.} We present \textbf{FlashTrie}, a GPU-native constrained decoding framework for generative retrieval that jointly optimizes constraint representation and decoding computation. Building on succinct tries, we redesign MARISA for integer-token vocabularies using a bit-compressed layout that reduces index size and keeps the constraint structure resident in GPU high-bandwidth memory (HBM). We further introduce GPU-friendly search and sort primitives, parallel trie child matching and heap-free beam selection, to replace pointer-heavy traversal and reduce divergence. Constrained beam search is executed fully on-device via a cooperative multi-step CUDA kernel that performs expansion, validation, and pruning without repeated kernel relaunch overhead across decoding steps. Two-level parallelism, beam-parallel and top-$K$-parallel execution, exploits hierarchical GPU concurrency, saturating hundreds of streaming multiprocessors (SMs) and maintaining high occupancy across the device.  FlashTrie  scales constrained decoding to billion-scale constraint libraries on GPU, a regime previously feasible only in offline settings, while achieving up to 24$\times$ speedup over optimized CPU baselines and reducing trie-search latency to under 3\,ms. In production A/B testing on a large-scale sponsored-search system, these latency gains enable wider beam search within strict serving budgets, translating to a +0.71\% revenue lift.

\section{Design and Implementation}
\label{sec:methods}
\label{subsec:setup}
We consider constrained decoding with a trie $\mathcal{T}$ over
vocabulary $\mathcal{V}$. FlashTrie applies to both AR and NAR
decoding; we focus on NAR in experiments because larger branching
factors make constraint checking more expensive. At decoding step $t$
the base model emits top-$K$
candidates $(x_{t,i}, \ell_{t,i})$ with $\ell_{t,i}=\log
p(x_{t,i}\mid x_{<t})$; trie-constrained beam search keeps the
top-$B$ valid prefixes under length-normalized
log-probability~\cite{wu2016googles}
(Equation~\ref{eq:length_norm},
Appendix~\ref{app:setup_detail}). For comparison, we use two CPU
baselines built on the same LOUDS+tail/link trie
skeleton~\cite{yata2011marisa} (Appendix~\ref{app:louds}).

\subsection{CPU Baselines: MARISA-Int and MARISA-Opt}
\label{subsec:mi}
\label{subsec:iom}
\textbf{MARISA-Int} extends open-source MARISA~\cite{yata2011marisa} from character-level keys to 32-bit integer tokens and adds a minimal beam-search layer, while preserving the original build pipeline. It runs single-threaded on CPU.
\textbf{MARISA-Opt} is a stronger CPU baseline: adds trie traversal with prefix-state caching, a multithreaded worker pool, two-pointer merge between sorted proposals and child labels ($O(K{+}\Delta)$ per beam), and a bounded min-heap for top-$B$ pruning ($O(C\log B)$). Design details and per-operation complexity are in Appendices~\ref{app:baselines} and~\ref{app:cpu_ops}.

\subsection{FlashTrie: GPU-Resident Constrained Beam Search}
\label{subsec:fm}

\begin{figure}[tbp]
  \centering
  \includegraphics[width=\columnwidth]{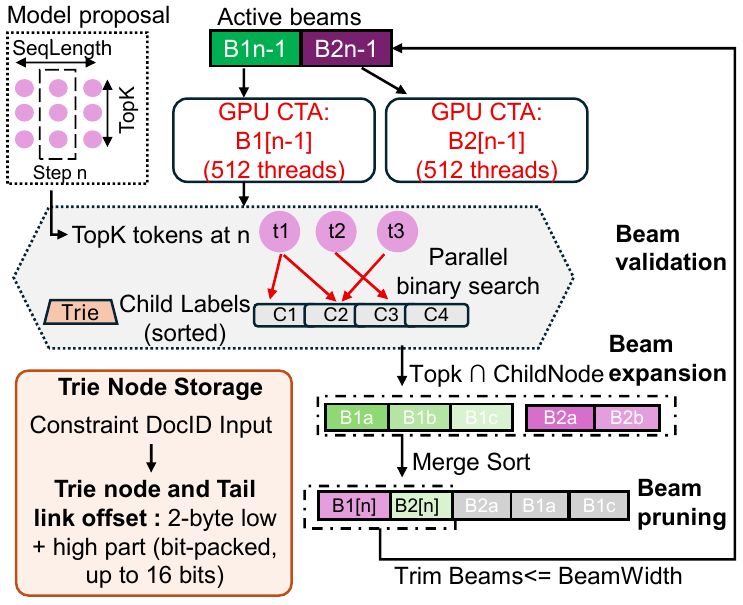}
  \caption{Overview of  FlashTrie pipeline. At each decoding step, active beam states are mapped to thread blocks (CTAs) and processed in parallel across the threads of a CTA for trie-based validation and beam expansion all executed on-device. (\textbf{Inset})  FlashTrie extends Narrow-LOUDS representation to integer tokens stored in compact structured arrays.}
  \label{fig:beam_method}
\end{figure}
FlashTrie redesigns both the trie data structure and the decoding algorithm for GPU execution. The result is a GPU-resident constraint index that preserves succinct-trie space efficiency while reducing memory stalls and exposing hierarchical parallelism.

\paragraph{Narrow LOUDS for integer tokens}
Published MARISA targets 8-bit alphabets. FlashTrie extends the layout to 32-bit integer tokens to enable large vocabularies (over a million)  but narrows the per-node slot by splitting both label and suffix index into a 2-byte low part in the base array and a bit-packed high part in a secondary array for nodes that need them. This halves the
dominant array, shrinks the constraint index, and
keeps billion-scale libraries in GPU HBM (Figure~\ref{fig:beam_method}). It also improves locality and reduces pointer-chasing stalls as fewer bits are fetched per access
(Section~\ref{subsec:build}).

\paragraph{Kernel Execution and Parallelism.}
FlashTrie executes constrained beam search in a single cooperative kernel, avoiding per-step CPU-GPU orchestration.  The GPU grid has several multiprocessors (SM)  
with each SM having one cooperative thread block (CTA)~\cite{nvidia_cuda_programming_guide,nvidia_a100_whitepaper}. Queries are batched across the SMs. To enable parallel beam search, within each query, an active beam is assigned to one CTA. A CTA has 512 threads that perform top-$K$ expansion and trie-based validation via batched child lookup. A grid-synchronized step then merges and prunes beams using parallel merge sort to enforce beam width. 

At each decoding step, FlashTrie exposes two levels of parallelism. First, a beam-parallel expansion across all $BW$ active beams. Second, proposal-parallel validation within each beam, where each thread tests one of the top-$K$ language-model proposals through binary search over sorted trie child labels. The full $T$-step loop runs in one GPU launch (Algorithm~\ref{alg:beamsearch}), hence at $BW = K = 1000$, this yields up to one million concurrent child-lookups across the GPU grid.

After expansion, the CPU heap is replaced by an append-only candidate buffer and a parallel top-$B$ selection kernel. This removes lock contention and pointer chasing while keeping intermediate candidates in fast on-chip memory (Appendix~\ref{app:gpu_detail}).

\begin{algorithm}
\small
\caption{\textsc{FlashTrie} beam search. One persistent GPU kernel
executes all $T$ steps on device, exposing beam-parallel expansion and
proposal-parallel validation. Kernel details:
Appendix~\ref{app:gpu_detail}.}
\label{alg:beamsearch}
\begin{algorithmic}[1]
  \Require trie $\mathcal{T}$, initial state $s_0$, beam width $BW$, length $T$,
           per-step LM proposals $\textsc{TopK}(s,K)$
  \State $\mathit{cur} \leftarrow \{s_0\}$
  \For{$t = 0$ \textbf{to} $T{-}1$}
    \State $\mathit{next} \leftarrow \emptyset$
    \For{each $s \in \mathit{cur}$ \textbf{in parallel}}
      \Comment{(i) beam-parallel}
      \State $(\tau_{1{:}K}, \rho_{1{:}K}) \leftarrow \textsc{TopK}(s, K)$
      \For{$k = 1$ \textbf{to} $K$ \textbf{in parallel}}
        \Comment{(ii) proposal-parallel}
        \State $u \leftarrow \textsc{ChildLookup}(\mathcal{T}, s.v, \tau_k)$
          \Comment{$O(\log\Delta)$ binary search}
        \If{$u\!\neq\!\bot$ \textbf{and} $s.\sigma{+}\rho_k>\theta_{\mathrm{sent}}$}
          \State append $(s,u,\rho_k)$ to $\mathit{next}$
            \Comment{lock-free}
        \EndIf
      \EndFor
    \EndFor
    \State \textbf{barrier};\ \ \ $\mathit{cur} \leftarrow \textsc{Top-}B(\mathit{next})$
      \Comment{on-device}
  \EndFor
  \State \Return \textsc{Backtrace}($\mathit{cur}$)
\end{algorithmic}
\end{algorithm}

\section{Experimental Setup}
\label{sec:methodology}
\label{subsec:setup_results}

We evaluate on 13,000 retrieval requests from a production NAR model with a 2.2M-token vocabulary. Each request generates top-$K$ proposals ($ \in {100, \ldots, 1000}$, beamwidth (BW) $= K$) over $T=8$ decoding steps. The constraint trie is built with 800M keyword sequences and is resident on a single A100 80GB GPU paired with an AMD EPYC 7V13 host. For MARISA-Opt, we use 8 CPU workers to saturate the 8-core allocation; full hardware isolation details in Appendix~\ref{app:setup_detail}.

We measure per-request latency (mean, p50/p90/p95/p99),
throughput under batched serving ($b \in {1,4,8,16,32}$ queries), on-disk
index size and wall-clock build time, and Precision@100/200
scored by a transformer teacher
(Section~\ref{subsec:precision}). Latency timing spans the full request path (C++ entry to result return, including GPU stream synchronization). All timings average 10 passes over the full 13,000-request set after one warm-up discard. Details in Appendix~\ref{app:setup_detail}.

\section{Results}
\label{sec:results}

  \subsection{Trie Build Time and Index Size}
  \label{subsec:build}

\begin{figure}[tbp]
    \centering
    \includegraphics[width=\columnwidth]{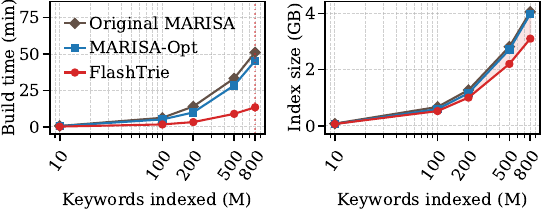}
    \caption{Trie construction scaling on 10M--800M keyword sequences, (\textbf{Left}) Build time, (\textbf{Right}) Constraint index size on disk; at runtime, the full trie is resident in GPU memory. At 10M keys, FlashTrie builds in 0.15\,min and
      uses 0.061\,GB, compared with 0.48\,min / 0.060\,GB for MARISA-Opt and
      0.58\,min / 0.075\,GB for MARISA-Int. At 800M keys, FlashTrie
      reaches 13.4\,min and 3.1\,GB, versus 45.3\,min / 4\,GB for
      MARISA-Opt and 51.1\,min / 4\,GB for MARISA-Int.}
    \label{fig:build_scaling}
  \end{figure}

  We begin by evaluating how FlashTrie's storage redesign affects two immediate construction outcomes: trie build time and on-disk index size.FlashTrie reduces both as the
  constraint set grows (Figure~\ref{fig:build_scaling}). Across 10M--800M sequences, the widening gap reflects a shift in the dominant cost: at small scales, fixed LOUDS and tail-construction overheads dominate (Appendix~\ref{app:louds}), whereas at larger scales the per-edge link machinery and traversal bookkeeping become the bottleneck
  (Appendix~\ref{app:cpu_ops}). MARISA-Int and MARISA-Opt remain close because they retain the same full-width node storage layout, while FlashTrie reduces per-node work by  narrowing link offsets and splitting each 32-bit label into
  a compact base field plus packed high bits (Section~\ref{subsec:fm}). As the corpus grows, those savings compound. Even at such a large constraint index as 800M, trie size is only around 3\,GB in FlashTrie,  allowing the trie to fit entirely in GPU HBM and enabling deployment on smaller GPUs like T4 or A100 MIG. Further, FlashTrie achieves
  3.4$\times$ faster build time than MARISA-Opt and uses 22\% less index space. Taken together, these construction-time gains establish the practical value of the storage redesign, which is one of the key enablers of the latency results that follow.

  \subsection{Latency Results}
  \label{subsec:latency_cpu}

\begin{figure}[tbp]
    \centering
    \includegraphics[width=\columnwidth]{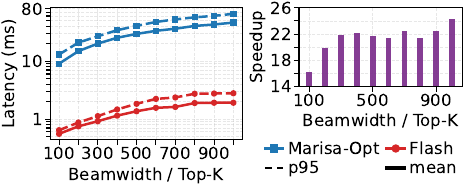}
    \caption{Per-request trie-search latency  and GPU speedup
      vs.\ beam width $BW$ at $b{=}1$ 
      ( $|\mathcal{V}|$=2.2M, 800M-key trie).
      \textbf{(Left)} Mean (solid) and p95 (dashed) latency. CPU mean
      rises from $9.0$\,ms ($BW{=}100$) to $46.3$\,ms ($BW{=}1000$),
      with p95 $65.5$\,ms and p99 $76.7$\,ms at the top end. GPU mean
      stays sub-$2$\,ms across the sweep ($0.55$--$1.91$\,ms), with
      p95 $\leq 2.79$\,ms and p99 $\leq 3.31$\,ms.
      \textbf{(Right)} Mean GPU speedup climbs from
      $ 16\times$ to $ 24\times$. Full per-percentile table:
      Appendix~\ref{app:latency_percentiles}.}
    \label{fig:latency_cpu_gpu}
  \end{figure}

  Next, we evaluate how GPU occupancy-focused kernel optimizations, together with full on-device residency of the constraint index, impact trie-search latency as beam width increases. On the GPU, each request runs in a single persistent cooperative
  kernel that executes every decoding step on-device and
  synchronises across steps with a grid-wide barrier
  (Section~\ref{subsec:fm}). The host pays launch overhead once per query rather than once per step, keeping latency in a narrow band as $BW$ grows (Figure~\ref{fig:latency_cpu_gpu}). GPU mean latency rises from $0.56$\,ms at $K{=}100$ to
  $1.91$\,ms at $K{=}1000$, with p95 below $2.79$\,ms and p99 below
  $3.31$\,ms. MARISA-Opt instead expands beams
  sequentially on a fixed-size worker pool
  (Section~\ref{subsec:iom}), and its mean
  latency rises from $9.03$\,ms to $46.30$\,ms, while its p99 rises from
  $15.08$\,ms to $76.71$\,ms over the same sweep. The resulting mean
  speedup grows from $16.3\times$ to $24.2\times$ as $BW/K$ increases (Figure~\ref{fig:latency_cpu_gpu}). We observe slight dips near $K{\approx}$ 500 and 800,
  where the on-device top-$B$ sorter changes thread capacity
  (Appendix~\ref{app:gpu_detail}).

  MARISA-Int (Section~\ref{subsec:mi}) is intractable at
  billion-keyword scale (${>}3$\,h for a single $K{=}1000$
  query on a 10-query subsample), so we adopt MARISA-Opt as the
  primary CPU comparator; full MARISA-Int measurements are in
  Appendix~\ref{app:marisa_int}.  Maintaining low latency as $BW$ increases shows that FlashTrie's optimizations remain effective at scale; next, we break down kernel runtime by phase to identify which components dominate as $BW$ grows.

  \subsection{Runtime Breakdown}
  \label{subsec:runtime_breakdown}
  
  To profile the latency trend, we instrument FlashTrie's cooperative kernel and
  measure GPU runtime across four phases (Figure~\ref{fig:beam_method}): beam expansion, validation, pruning and grid
  sync. Figure~\ref{fig:throughput} (left) shows the dominant cost shifts with beam width: at $K{=}100$, expansion
 dominates ($\approx$66\%) as trie traversal is the bottleneck. As $BW$ grows,
  expansion remains inexpensive because parallel beams share upper trie nodes. In contrast,
  the number of surviving candidates scales with $B{\times}K$,
  increasing validation and pruning cost; at $BW{=}1000$, validation
  becomes dominant ($\approx$43\% of total runtime). Pruning remains stable at
  $15$--$19\%$, while sync overhead stays small
  ($\le 6\%$, decreasing to $3.5\%$), confirming that the
  persistent kernel avoids per-step host orchestration with
  negligible barrier overhead. Measurement details are in Appendix~\ref{app:runtime_breakdown}. Overall, the bottleneck shifts from trie traversal to candidate processing at larger BW. We next quantify the system-level impact using batch-throughput measurements.

  \subsection{Throughput Scaling with Batch Size}
  \label{subsec:throughput}

\begin{figure}[!t]
    \noindent
    \includegraphics[width=\columnwidth,height=0.4\textheight,keepaspectratio]{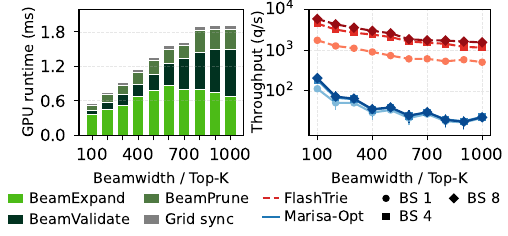}
    \caption{(\textbf{Left}) GPU  runtime breakdown of FlashTrie's
cooperative beam-search kernel vs.\ beam width $BW$ ($b{=}1$,
averaged over 13{,}000 requests).  As $BW$ grows, the bottleneck shifts from
     trie traversal (beam expansion) to candidate handling (beam validation). (\textbf{Right}) Throughput (mean q/s) vs.\
      $BW$, per batch
      size. FlashTrie peaks at $b{=}8$ for all $BW$
      ($6{,}623$\,q/s at $BW{=}100$, $1{,}621$\,q/s at $BW{=}1000$),
      while MARISA-Opt flattens earlier. At each
      system's own best batch size, FlashTrie delivers
      $32\times$--$71\times$ more throughput than MARISA-Opt.}
    \label{fig:throughput}
  \end{figure}

  Latency improvements are operationally meaningful if they convert into higher serving capacity under realistic batching. In Figure~\ref{fig:throughput} (right), the throughput curves reflect different batching regimes. FlashTrie's throughput rises steeply from $b{=}1$ to $b{=}8$ as
  batching fills more GPU execution resources, peaking at
   $1{,}621$\,q/s for $BW{=}1000$. This is $70\times$ higher than MARISA-Opt at its own peak batch size. MARISA-Opt scales only modestly with batch size because its CPU thread pool is quickly saturated: once all workers are occupied, extra queries mostly queue (Section~\ref{subsec:iom}). FlashTrie benefits more from batching because each query is mapped to a GPU block with many threads, so larger batches better fill the device's execution resources. Beyond $b{=}8$, however, thread synchronization and partition overhead start to dominate, so throughput falls. These throughput gains indicate more effective GPU utilization under batched serving, with higher occupancy and better device saturation as workload increases.
  
  \subsection{Retrieval Quality}
  \label{subsec:precision}

  Given that FlashTrie maintains only a fraction of MARISA-Opt's latency even at larger beam widths, we next verify that this highly optimized decoding path does not compromise retrieval quality. Table~\ref{tab:precision} shows that retrieval quality improves with beam width. P@100 increases from 0.53 at $BW{=}100$ to 0.78 at $BW{=}1000$, and P@200 increases from 0.52 to 0.69. Gains taper after $BW{\approx}700$, but the trend remains monotonic overall at the operating scale of interest.

  \begin{table}[tbp]
    \centering
    \caption{Retrieval precision at cutoffs 100 and 200 on the 13K-request dataset, scored by a Transformer cross-encoder~\cite{valluri2025pixnar}. MARISA-Opt and FlashTrie agree within $0.001$ for every $BW$ (collapsed into one column). Prec@200 at $BW{=}100$ equals Prec@100 due to a truncated denominator}
    \label{tab:precision}
    \footnotesize
    \setlength{\tabcolsep}{3pt}
    \begin{tabular}{l|*{10}{c}}
       \toprule
      \textbf{$BW$}   & 100 & 200 & 300 & 400 & 500 & 600 & 700 & 800 & 900 & 1000 \\
      \midrule
      \textbf{P100} & .53 & .68 & .72 & .74 & .75 & .76 & .77 & .77 & .78 & .78 \\
      \textbf{P200} & .52 & .46 & .56 & .61 & .64 & .65 & .67 & .66 & .68 & .69 \\
      \bottomrule
    \end{tabular}
  \end{table}

   FlashTrie matches MARISA-Opt to within $0.001$ across all settings, indicating that GPU execution remains in parity with the CPU implementation. More importantly, in latency-critical serving (e.g., sponsored search; Section~\ref{subsec:online_flight}), MARISA-Opt becomes impractical beyond roughly $BW{\ge}200$ as beam width grows. Thus, the quality gains from larger beam widths are operationally realizable only with FlashTrie, which keeps constrained decoding within tight online latency budgets where large-$BW$ MARISA-Opt cannot.

  \subsection{Ablations: Isolating  Trie and Search-Algorithm
    Contributions}
  \label{subsec:ablations}

   Section~\ref{subsec:latency_cpu} combines three effects: execution substrate (CPU vs.\ GPU), trie-based constraint checking, and inner-loop child-search algorithm.  We isolate the latter two with GPU-resident ablations that share FlashTrie's cooperative kernel, scoring rules, thresholds, and length normalization. 
   
  Per-Position Token (PPT)
   is a depth-only filter, not a prefix-constrained decoder: at step $t$, it intersects top-$K$ model tokens with a global sorted token list for depth $t$, then returns all matches regardless of beam prefix. Because this depth-level token set is shared, the matched set is computed once and reused across beams.
   
  Linear-probe keeps
  the LOUDS trie unchanged but swaps binary search for a shared-memory linear scan over children,
  isolating $\mathcal{O}(\Delta)$ vs.\ $\mathcal{O}(\log\Delta)$ behavior.
  Setup details 
  are in Appendices~\ref{app:ppt_details}
  and~\ref{app:linear_ablation}.

  \begin{table}[tbp]
  \centering
  \caption{Per-request latency (ms) for two GPU ablations vs.\
    FlashTrie at $b{=}1$.  PPT removes trie constraints; Linear-probe keeps the trie but replaces binary search with linear scan in the child range.
    We report p95 speedup as $\text{ablation p95} / \text{FlashTrie p95}$. Every other $K$ is shown; full sweep in Appendix~\ref{app:ablations_full}, Table~\ref{tab:ablations_full}.}
  \label{tab:ablations}
  \small
  \setlength{\tabcolsep}{3pt}
  \begin{tabular}{c rr c rr c}
     \toprule
    & \multicolumn{3}{c}{\textbf{PPT on GPU}}
    & \multicolumn{3}{c}{\textbf{Linear-probe on GPU}} \\
    \cmidrule(lr){2-4} \cmidrule(lr){5-7}
      \textbf{$BW$}
         & Mean & p95 & \textbf{Sp.}
         & Mean & p95 & \textbf{Sp.} \\
       \midrule
        100  & 0.57  & 0.79  & $1.2\times$ & 110.2 & 134.1 & $209\times$ \\
        300  & 1.90  & 2.69  & $2.4\times$ & 123.9 & 149.1 & $133\times$ \\
        500  & 3.54  & 5.20  & $2.8\times$ & 128.6 & 155.7 & $85\times$  \\
        700  & 6.98  & 10.30 & $4.4\times$ & 141.1 & 178.8 & $76\times$  \\
        900  & 9.73  & 14.36 & $5.2\times$ & 151.2 & 194.3 & $71\times$  \\
       1000  & 11.28 & 16.65 & $6.0\times$ & 154.3 & 198.7 & $71\times$  \\
       \bottomrule
  \end{tabular}
\end{table}

  Table~\ref{tab:ablations} shows three takeaways. (i) A permissive GPU baseline (PPT) is fast but not constraint-equivalent. (ii) adding prefix-constrained trie checking (PPT to FlashTrie) provides an additional gain that grows
  with  $BW$ (from $1.2\times$ to $6.0\times$ in p95). (iii) binary search over child labels is critical (Linear-probe to FlashTrie), with $71\times$--$209\times$ p95 degradation when replaced by linear scan. These ablations show that both prefix-constrained trie filtering and logarithmic child lookup are essential, indicating the gains come from core algorithmic design rather than hardware alone.

  \subsection{Public-Dataset Reproducibility}
  \label{subsec:public_data}

  To reproduce the latency claim on public artifacts, we
  synthesise a parallel workload from NQ-Open
  queries~\cite{kwiatkowski2019natural,lee2019nqopen}, the
  GENRE-KILT BART-large
  checkpoint~\cite{decao2021genre,lewis2020bart}, and the
  $\sim$6M KILT Wikipedia titles~\cite{petroni2021kilt} as the
  constraint library. We sweep beam width
  $BW \in \{100, 200, \ldots, 1000\}$ over all 3{,}600 NQ-Open
  validation queries ($T{=}16$ decoding positions) and measure
  per-request trie-search latency on the same hardware as
  Section~\ref{subsec:setup_results}.

\begin{figure}[tbp]
    \centering
    \includegraphics[width=\columnwidth]{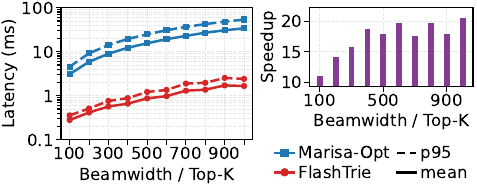}
    \caption{Latency on the NQ+GENRE public workload
      (3{,}600 queries, $T{=}16$, ${\sim}6$M-title trie) vs.\ beam
      width $BW$. GPU mean latency rises from $0.28$\,ms ($BW{=}100$)
      to $1.67$\,ms ($BW{=}1000$); MARISA-Opt mean grows from
      $3.11$\,ms to $34.23$\,ms. Mean speedup: $11.0\times$ to
      $20.5\times$.}
    \label{fig:nq_genre_speedup}
  \end{figure}

  Figure~\ref{fig:nq_genre_speedup} shows that FlashTrie
  reproduces the internal-dataset speedup trend
  (Figure~\ref{fig:latency_cpu_gpu}) on entirely public
  artifacts. The $p99$ speedup increases from $12.8\times$ to $22.7\times$, indicating both lower and more predictable tail latency across the full $BW$-sweep. A detailed noise-injection ablation studying
  the effect of cross-position incoherence (NAR-style proposal
  grids) is in Appendix~\ref{app:nq_genre_setup}. Reproducing the same scaling pattern on public data supports generality beyond internal traffic, leading to the final test on live production impact.

  \subsection{Online A/B Testing}
  \label{subsec:online_flight}
  We validate our offline findings with a randomized online A/B experiment on live traffic from a popular commercial search engine. The goal is to test whether the larger beam width enabled by FlashTrie yields meaningful online gains in user engagement and revenue, i.e., whether offline  improvements transfer to gains in production. The experiment runs for 16 consecutive days across multiple countries, with English and non-English results reported separately in Table~\ref{tab:online_flight}.

  Both arms deploy the same NAR generative retrieval model~\cite{valluri2025pixnar} on an A100 MIG 10\,GB slice. In control, production MARISA-Opt at $BW{=}200$ is already near the online limit ($\approx$25\,ms p95; $BW{>}200$ violates latency constraints). The treatment swaps MARISA-Opt trie stage for FlashTrie and uses the recovered latency headroom to increase beam width to $BW_{\text{trt}}{=}600$, while keeping the model and downstream ranking stack unchanged. FlashTrie sustains $BW{=}600$ at $15/17$\,ms (p50/p95), exploring over $2\times$ more candidates within the same serving budget.

 The treatment yields statistically significant gains,
  including a $0.71\%$ lift in revenue, without degradation in ad quality. We further analyze key metrics ad coverage, impressions, clicks and  defect rate. Ad defect, measured using offline relevance models, denotes the proportion of irrelevant ads shown to users. FlashTrie increases user engagement, with statistically significant click improvements of $0.17\%$ for English queries and $0.20\%$ for non-English queries. Since the underlying model and its top-K proposals are the same, these gains result mainly from increasing beam width, enabling retrieval of more candidate ads and expanding the proposal set passed to downstream ranking.

  \begin{table}[ht]
    \centering
    \caption{Online A/B test ($\Delta$) on a commercial search engine for FlashTrie vs.\ 
      CPU MARISA-Opt, reported by language segment.
      All metrics are statistically significant ($\alpha{=}0.05$). See Appendix~\ref{app:online_flight_protocol} for the full protocol.}
    \label{tab:online_flight}
    \scriptsize
    \setlength{\tabcolsep}{1pt}
    \resizebox{\columnwidth}{!}{%
    \begin{tabular}{@{}lccccc@{}}
       \toprule
      \textbf{Language } & \textbf{Revenue} &
      \textbf{Coverage} & \textbf{Impression} &
      \textbf{Clicks} & \textbf{Latency} \\
      \midrule
      English      & $+0.71\%$ & $+0.22\%$ & $+0.21\%$ & $+0.17\%$ & $-32\%$ \\
      Non-EN  & $+0.68\%$ & $+0.25\%$ & $+0.37\%$ & $+0.20\%$ & $-32\%$ \\
      \bottomrule
    \end{tabular}%
    }
  \end{table}

  Ad coverage and impressions increase as wider beams recover monetizable candidates that narrower beams discard, while revenue gains suggest these candidates correspond to higher-value matches. Importantly, defect rates are nonsignificant within $\pm1\%$, indicating no degradation in result quality. Overall, the latency headroom enabled by FlashTrie allows a substantial expansion of beam width, translating directly into improved system performance.

  \section*{Discussion}
  \label{sec:discussion}
  FlashTrie exemplifies  a broader accelerator-era pattern: data structures traditionally implemented on CPUs can become effective on GPUs when their access patterns are redesigned for parallel execution.  Our central finding is that constrained decoding need not remain a CPU bottleneck. A static succinct trie can be made GPU-native when lookup and pruning are structured for batched execution and high device utilization. This reduces decoding latency even at wider beam widths, translating to lower end-to-end inference latency under strict serving constraints.

  This shift is important because retrieval quality in generative systems improves at larger beam widths. In our online setting, FlashTrie enables wider beams within strict latency budgets and converts that headroom into measurable gains in user engagement and revenue without degrading ad quality. More broadly, FlashTrie applies to constrained generation tasks with large static vocabularies, such as retrieval IDs, product catalogs, or entity linking, that can benefit from accelerator-resident constraint enforcement that is both fast and scalable.

  \section{Limitations}
  \label{sec:limitations}

  The current implementation requires the full constraint trie
  to reside within a single GPU's VRAM. At 800M keywords the
  trie occupies $3.1$\,GB on the A100 80\,GB, leaving ample
  headroom for co-located weights; for larger libraries a
  natural extension is to shard subtrees across GPUs over
  NVLink/NVSwitch and aggregate candidates per step.

  FlashTrie's batched scheduler allocates a fixed SM
  partition to each query (Section~\ref{subsec:throughput}),
  avoiding cross-slot synchronisation but capping multi-query
  parallelism, dynamic or work-stealing alternatives that
  preserve persistent-kernel coherence are a natural next step.

  All experiments use an NVIDIA A100 80\,GB PCIe GPU. The
  persistent-kernel and cooperative-groups patterns are
  supported on all CUDA Compute Capability $\geq 7.0$ devices,
  but quantitative characterisation on H100/B100 remains open.

\bibliography{example_paper}

\appendix
\begin{center}
\textbf{Appendix}
\end{center}

\section{Trie Background: LOUDS with Tail/Link Compression}
\label{app:louds}

All three systems share the same LOUDS backbone~\cite{yata2011marisa,jacobson1989succinct}: a
$2n{+}1$-bit topology vector with constant-time \texttt{rank}/\texttt{select},
plus per-node \texttt{bases\_} labels and \texttt{terminal\_flags\_}. What
differs across systems is not the core topology but auxiliary build-time
structures (Patricia/TAIL, link flags, and cache) and the search algorithm
used on top of this layout (Table~\ref{tab:three_system_contrast}).

\paragraph{Patricia single-child compression and the TAIL buffer.}
Patricia compression~\cite{morrison1968patricia} folds maximal single-child
chains into a head node plus an offset into \texttt{tail\_}; link nodes store
offsets, not labels. Without suffix sharing, this is usually a loss for wide
labels. MARISA's trie-of-tails variant implements this same idea with a
recursive TAIL buffer~\cite{yata2011marisa}. The net benefit depends on suffix
deduplication, which can be summarized by:
\[
\text{folded}(k,w,\alpha)
\;\approx\; w + \alpha\, k w + O(\log |\texttt{tail\_}|).
\]
Folding beats expansion iff $\alpha\, kw + w < kw$, i.e.,
\begin{equation}
\boxed{\;\alpha \;<\; 1 - \tfrac{1}{k}\,.\;}
\label{eq:patricia_breakeven}
\end{equation}
In byte alphabets this criterion often holds; in wide 32-bit token alphabets
suffix sharing is much weaker and the supporting metadata becomes relatively
expensive. This is why MARISA-Opt disables Patricia/TAIL in our integer-token
setting (Section~\ref{subsec:iom}). Recursive TAIL (\texttt{num\_tries}) helps
only when the same criterion holds at each level~\cite{yata2011marisa}; we use
\texttt{num\_tries}=1.

\paragraph{The \texttt{link\_flags\_} bit-vector and \texttt{cache\_}.}
\texttt{link\_flags\_} tags whether \texttt{bases\_} holds a label or a TAIL
offset; its rank index and high-bit arrays scale with the number of link nodes
in MARISA's succinct-trie layout~\cite{yata2011marisa}. When Patricia is
disabled, these structures largely collapse. MARISA's optional
\texttt{cache\_} prefetch table speeds common top-of-trie transitions but adds
build cost and index bytes~\cite{yata2011marisa}. These component-level
tradeoffs are summarized in Table~\ref{tab:three_system_contrast}; for full
implementation details, see \citet{yata2011marisa}.

\section{CPU Baselines: Full Design Details and System Contrast}
\label{app:baselines}

This appendix gives the full design description for the two CPU
baselines summarised in Section~\ref{subsec:mi} and the
axis-by-axis contrast between all three systems
(Table~\ref{tab:three_system_contrast}).

\subsection{MARISA-Int}
\label{app:baselines:mi}

MARISA-Int extends the open-source MARISA codebase~\cite{yata2011marisa}
with 32-bit integer-token sequences, reusing the original character-oriented storage layout and traversal routines without modifying the build pipeline.
Since upstream MARISA provides only single-sequence operations (e.g., \texttt{lookup}, \texttt{predictive\_search}), MARISA-Int adds a minimal beam-search layer. Each candidate is evaluated by repeatedly invoking these routines on the full prefix from the root, performing up to two root-to-node traversals per (beam, proposal) pair. No trie state is reused across steps, and pruning is implemented via \texttt{std::partial\_sort} over a flat vector. The search is single-threaded, and all operations execute on the host CPU.

\subsection{MARISA-Opt}
\label{app:baselines:iom}

MARISA-Opt builds on MARISA-Int while retaining the LOUDS-based trie layout and on-disk format, but introduces a set of implementation optimizations that make it a competitive CPU baseline. These include compact link-offset encoding, stateful traversal, multi-threaded execution, efficient per-beam expansion, and incremental top-$B$ pruning (full per-operation derivations in Appendix~\ref{app:cpu_ops}). Comparing FlashTrie against this optimized baseline ensures that performance gains reflect on-device parallelism rather than unoptimised CPU execution.

MARISA-Opt reduces index size by splitting link offsets across the per-node 32-bit slot and a secondary array, storing only the low-order 8 bits in the slot and packing the remaining bits compactly. The secondary array is allocated only when needed, avoiding unnecessary overhead. Token labels themselves remain in full in the per-node slot; only the link offsets are split. On CPU, MARISA-Opt does not narrow the per-node slot itself, the slot is still 32 bits, so at each link node 24 of those 32 bits are now unused padding.

For integer-token alphabets, MARISA-Opt further removes structures whose benefit is specific to byte-alphabet inputs (Appendix~\ref{app:louds}). First, Patricia single-child link compression is disabled in the build path: every input token becomes its own LOUDS node, no link nodes are emitted, and the recursive trie-of-tails terminates after the first level regardless of the configured \texttt{num\_tries}. As a consequence, \texttt{extras\_}, the TAIL buffer, and the rank/select index over \texttt{link\_flags\_} are all empty in the on-disk index, and the link-offset split encoding described above is exercised only by character workloads. Second, the build-time prefetch cache table (\texttt{cache\_}, used to amortize top-of-trie descents at lookup time) is skipped at construction and is absent from the on-disk index. Third, the in-process multikey quicksort over \texttt{Vector<Key>} is replaced by an external \texttt{sort -u -V} Unix pipeline that delivers unique, lexicographically ordered token sequences to the builder, so duplicate keys are removed before any trie work begins.

The combined effect is that on a 32-bit alphabet, where each TAIL token would still occupy 4 bytes and each TAIL reference would carry non-trivial \texttt{extras\_} and \texttt{link\_flags\_} overhead, the bytes that Patricia compression would remove from \texttt{bases\_} are smaller than the bytes its supporting structures would add. Disabling the compression therefore shrinks the on-disk index and shortens construction simultaneously; this is the dominant driver of MARISA-Opt's index-size and build-time advantage over MARISA-Int in Section~\ref{subsec:build}.

Unlike MARISA-Int, which performs repeated root-to-node traversals at each decoding step, MARISA-Opt propagates trie state across beams. Each extension requires a single LOUDS navigation step from the cached node, making per-candidate cost independent of prefix length.

Work is parallelized across beams using a fixed-size CPU worker pool, while per-beam expansion is performed using a linear-time two-pointer merge between sorted proposals and sorted trie children. This reduces matching cost from $O(K\Delta)$ to $O(K{+}\Delta)$ with sequential memory access. Threshold-based pruning is applied during expansion to discard low-probability candidates early.

Instead of end-of-step sorting, MARISA-Opt maintains a shared bounded min-heap of size $BW$, reducing pruning cost from $O(C\log C)$ to $O(C\log BW)$, where $C$ is the number of candidates generated per step and $BW$ is the beam width, while keeping the top-$B$ beams continuously updated.

\subsection{Axis-by-Axis System Contrast}
\label{app:baselines:table}

\begin{table*}[t]
\centering
\caption{Design and implementation contrast across the three systems.
All three share the LOUDS$+$tail/link trie skeleton
(Appendix~\ref{app:louds}); they differ in build-time storage
choices and in where/how the beam search is executed. The
\emph{Improves} column tags whether each axis primarily affects
search latency
(\tagLAT),
build time
(\tagBUILD),
or trie-on-disk size
(\tagSIZE).
Per-operation time complexity is reported separately in
Table~\ref{tab:complexity}. Background on Patricia single-child
compression, the TAIL buffer, the \texttt{link\_flags\_}
bit-vector, and the prefetch \texttt{cache\_} table referenced in
the rows below is given in Appendix~\ref{app:louds}.}
\label{tab:three_system_contrast}
\scriptsize
\setlength{\tabcolsep}{2pt}
\renewcommand{\arraystretch}{1.08}
\newcolumntype{Y}{>{\raggedright\arraybackslash}X}
\newcolumntype{Z}{>{\centering\arraybackslash}p{0.095\textwidth}}
\begin{tabularx}{\textwidth}{@{}>{\raggedright\arraybackslash}p{0.125\textwidth}YYYZ@{}}
 \toprule
\textbf{Aspect} & \textbf{MARISA-Int} & \textbf{MARISA-Opt} & \textbf{FlashTrie} & \textbf{Improves} \\
\midrule
\texttt{bases\_} slot type
  & \texttt{Vector<UInt32>}, 4\,B/entry
  & \texttt{Vector<UInt32>}, 4\,B/entry
  & \texttt{Vector<UInt16>}, 2\,B/entry
  & \tagSIZE \\
\addlinespace[1pt]
Label storage at non-link nodes
  & full 32-bit token in \texttt{bases\_}
  & full 32-bit token in \texttt{bases\_}; label not split
  & low 16 bits in \texttt{bases\_}; high bits in \texttt{extras\_} via \texttt{split\_label}
  & \tagSIZE \\
\addlinespace[1pt]
Link-offset storage at link nodes
  & full pointer in \texttt{bases\_};
  & full pointer in \texttt{bases\_}
  & low 16 bits in \texttt{bases\_}; high bits packed in \texttt{link\_extras\_}; 
  & \tagSIZE \\
\addlinespace[1pt]
\texttt{num\_tries} (trie-of-tails recursion depth)
  & 1--7, default 1
  & 1--7, default 1
  & 1 only 
  & \tagBUILD\newline\tagSIZE \\
\addlinespace[1pt]
Recursive TAIL block (\texttt{tail\_})
  & built; tokens stored uncompressed as \texttt{uint32}, one offset entry per link node
  & empty (no link nodes produced, so the recursive \texttt{build\_next\_trie} returns immediately and \texttt{tail\_} is never populated)
  & built; tokens stored as \texttt{uint32}, references compressed via bit-packed \texttt{link\_extras\_}
  & \tagBUILD\newline\tagSIZE \\
\addlinespace[1pt]
\texttt{link\_flags\_} rank/select index
  & full bit-vector with rank/select over true link positions
  & identically zero across all nodes; rank/select index collapses to constant overhead
  & full bit-vector with rank/select over true link positions
  & \tagSIZE \\
\addlinespace[1pt]
Build-time prefetch cache (\texttt{cache\_} table)
  & built; size set by \texttt{cache\_level}, persisted to disk to amortize top-of-trie descents at lookup time
  & skipped (\texttt{reserve\_cache}, per-node \texttt{cache<T>}, and \texttt{fill\_cache} all bypassed)
  & built
  & \tagBUILD\newline\tagSIZE \\
\addlinespace[1pt]
Build-time key sort and dedup
  & in-process multikey quicksort over \texttt{Vector<Key>}; duplicate inputs not removed
  & in-process multikey quicksort over \texttt{Vector<Key>}; deduplication done outside 
  & in-process multikey quicksort over \texttt{Vector<Key>}; deduplication done outside
  & \tagBUILD \\
\addlinespace[1pt]
Beam-search residency
  & host RAM
  & host RAM
  & GPU HBM (CUDA Unified Memory, eager-prefetched)
  & \tagLAT \\
\addlinespace[1pt]
Trie cursor primitive in beam search
  & \texttt{lookup} then \texttt{predictive\_search} from root; up to two full root-to-node walks per candidate, no state carried across steps
  & \texttt{predictive\_find\_children}: one LOUDS step from a cached \texttt{node\_id}
  & \texttt{extend\_\allowbreak beam\_\allowbreak binary\_\allowbreak child\_\allowbreak search}: one GPU LOUDS step; state in device buffers
  & \tagLAT \\
\addlinespace[1pt]
Parallelism axis
  & none (single thread serialises both beams and proposals)
  & across parent beams (proposals serial within a beam)
  & across parent beams and across proposals within a beam
  & \tagLAT \\
\addlinespace[1pt]
Workers / dispatch unit
  & 1 CPU thread
  & 8 \texttt{std::thread} workers; one parent beam per pool job
  & $\lfloor 108/b\rfloor$ CTAs $\times$ 512 threads/CTA per slot
  & \tagLAT \\
\addlinespace[1pt]
Per-beam inner loop
  & linear scan over $BW$ proposals; trie re-walked per candidate
  & two-pointer merge between sorted child labels and sorted top-$B$ proposals
  & one thread per proposal; parallel binary search against the sorted child-label array
  & \tagLAT \\
\addlinespace[1pt]
Top-$B$ pruning structure
  & \texttt{std::partial\_sort} over a flat \texttt{std::vector} of beam states
  & mutex-protected \texttt{std::priority\_queue} bounded heap
  & on-device block-merge-sort; registers/shared memory
  & \tagLAT \\
\addlinespace[1pt]
Batched-call mechanism
  & sequential per query
  & shared FIFO of (query, beam) jobs across 8 workers; step barrier
  & $b$ cooperative kernels on $b$ CUDA streams; disjoint SM partitions; no cross-slot sync
  & \tagLAT \\
\bottomrule
\end{tabularx}
\end{table*}

\section{CPU Trie Operations}
\label{app:cpu_ops}

The CPU trie operations used by MARISA-Int and MARISA-Opt are
inherited from the MARISA framework and are not contributions of
this work; we describe them in prose for completeness so that the
GPU adaptations in Appendix~\ref{app:gpu_detail} have a precise
reference point.

\paragraph{Child enumeration.}
Locating the children of a node $v$ is two rank/select operations
on \texttt{louds\_}. The position of $v$'s first child within
\texttt{louds\_} is $\operatorname{select}_0(v) + 1$, and the
corresponding child node index is
$\operatorname{select}_0(v) + 1 - v - 1$ (subtracting the LOUDS
header bits seen so far). The number of children
$\Delta$ is the distance from that position to the next $0$-bit;
on CPU this is a sequential scan over \texttt{louds\_}, and on GPU
it is replaced by \textsc{WarpNextUnset}
(Appendix~\ref{app:gpu_detail}). Labels of the $\Delta$ children
sit contiguously at \texttt{bases\_[base\_node..base\_node+$\Delta$)}
in sorted order.

\paragraph{\textsc{FindChild}.}
Given a parent $v$ and a target token $\tau$, MARISA does a binary
search over the $\Delta$ child labels. The only subtlety is that
when a child $u$ is a link node
($\texttt{link\_flags\_}[u] = 1$), its \texttt{bases\_} slot holds
a TAIL offset rather than a token, so the binary-search comparison
must peek into \texttt{tail\_} at that offset to recover the
first label of the suffix; the comparison itself is then ordinary
integer comparison. This costs one extra memory indirection per
visited link node and is the reason TAIL traversal must read at
most one tail byte per binary-search step rather than the entire
suffix. Per-call cost is $O(\log \Delta)$ comparisons, each $O(1)$
amortized.

\paragraph{\textsc{Lookup} and predictive search.}
A whole-key lookup is $L$ successive \textsc{FindChild} calls
followed by a check that the resulting node carries a terminal
flag; the key identifier is then
$\operatorname{rank}_1(\texttt{terminal\_flags\_}, v) - 1$. Common
prefix search and predictive search extend this loop with an
additional in-order walk of the subtree rooted at the deepest
matching node, paying $O(R)$ for $R$ returned keys on top of the
$O(L \log \Delta)$ descent. All of these operations execute
serially on CPU and are the per-beam unit of work that
FlashTrie's expansion kernel parallelises in
Appendix~\ref{app:gpu_detail}.

\section{GPU Implementation Details}
\label{app:gpu_detail}

FlashTrie replaces the serial \textsc{FindChild}/\textsc{Lookup}
loop of Appendix~\ref{app:cpu_ops} with a cooperative-kernel
pipeline that keeps the trie, the beam state, and the top-$K$
output resident on the device for the entire query~\cite{nvidia_cuda_programming_guide,nvidia_a100_whitepaper}.
This appendix documents the pieces that genuinely differ from the CPU baseline
and are not adequately specified by prose alone: tail-suffix
traversal, warp-cooperative LOUDS scanning, and large-$BW$ top-$K$
selection. Bookkeeping aspects, memory residency, output
serialization, and per-batch allocation, are described in prose
since they involve no concurrency subtleties.
Table~\ref{tab:complexity} summarises per-operation time
complexity for the GPU primitives used below alongside the CPU
counterparts; the $1/P$ factor is the per-thread reduction the GPU
contributes, with $P{=}512$ for per-CTA expansion and the full
grid for sort passes.

\begin{table}[tbp]
\centering
\caption{Per-operation time complexity. $L$ = key length, $Q$ = query
length, $R$ = result count, $\Delta$ = node out-degree, $T$ =
decoding steps, $BW$ = beam width, $BW$ = top-$K$ proposals per step,
$P$ = GPU parallelism, BS = Beam Search.}
\label{tab:complexity}
\begin{tabularx}{\columnwidth}{@{}lX@{}}
 \toprule
Operation & Time \\
\midrule
$\operatorname{rank}_1(i)$           & $O(1)$ \\
$\operatorname{select}_b(k)$         & $O(1)$ \\
\textsc{WarpNextUnset}               & $O(\Delta/32)$ \\
Lookup                               & $O(L\log\Delta)$ \\
Common prefix search                 & $O(Q)$ total \\
Predictive search                    & $O(Q+R)$ \\
MARISA-Opt BS            & $O(T{\cdot}B{\cdot}(K+\Delta))$ \\
FlashTrie BS           & $O\!\left(T\tfrac{BK}{P}\log\Delta+T\tfrac{B\log B}{P}\right)$ \\
MergeSort top-$B$                    & $O(B\log B/P)$ \\
\bottomrule
\end{tabularx}
\end{table}

\paragraph{Memory architecture.}
GPU memory is partitioned into three classes. The
\emph{static trie} (LOUDS bit-vectors, label arrays, tail buffer,
rank/select indices) lives in CUDA Unified Memory and is
prefetched on demand; this lets a single index back many concurrent
search slots without per-slot duplication. \emph{Per-batch search
state}, three frontier arenas, a backtrace arena for parent
pointers, and a beam-history buffer used during output
reconstruction, is device-resident, allocated once at
\texttt{init\_tbs\_gpu} time, and reused across requests so that no
\texttt{cudaMalloc} appears on the hot path. \emph{Per-request I/O}
(input top-$K$ tokens and log-probabilities, output beam sequences
and normalised scores) is held in pinned host buffers paired with
device mirrors, enabling asynchronous DMA that overlaps with kernel
execution.

\subsection{Tail Traversal}
\label{subsec:tail_traversal}

Beam expansion on the GPU runs in two passes per decoding step,
both driven by the cooperative kernel in
Algorithm~\ref{alg:beamsearch}. The first pass handles beams that
are already \emph{inside} a tail suffix (\texttt{s.in\_link}): such
a beam has exactly one valid continuation, the next token along
the tail chain, so the work is to (i) read that token from
\texttt{tail\_} at \texttt{s.link\_offset}, (ii) scan the top-$K$
proposals for a match, (iii) apply token- and sentence-level
log-prob thresholds, and (iv) emit the resulting beam either to
\texttt{selected} (if the proposal closes the chain on a terminal
node) or to \texttt{next}. Because the continuation is unique, no
binary search is needed and each beam costs $O(K/P)$ work across
the CTA's $P$ threads; this is \textsc{ExtendInLink} in
Algorithm~\ref{alg:extendinlink}.

The second pass handles beams sitting on ordinary LOUDS nodes. For
each such beam the kernel first computes the child range
$[\mathit{base\_node}, \mathit{base\_node}{+}\Delta)$ using
$\operatorname{select}_0$ and \textsc{WarpNextUnset}, then has each
thread test one top-$K$ proposal in parallel by binary-searching
the sorted child labels. Each comparison transparently peeks into
\texttt{tail\_} when the visited child is a link node, exactly as
in CPU \textsc{FindChild}. Surviving proposals become new beams
that are routed to \texttt{selected} or \texttt{next} on the same
terminal-/link-status rules as the first pass; this is
\textsc{ExtendNode} in Algorithm~\ref{alg:extendnode}. Splitting
the work into two passes keeps the warps that are walking tail
chains from interfering with the warps that are doing wider
parallel child search, which keeps lane utilisation high in both
modes.

\begin{algorithm}
\caption{ExtendInLink (device; one CTA per parent beam)}
\label{alg:extendinlink}
\begin{algorithmic}[1]
  \State $\tau_{\mathrm{next}},\,\mathit{end}
         \leftarrow\mathit{tail}.\operatorname{next\_token}(s.\mathit{link\_offset})$
  \For{$k=\mathit{threadIdx.x}$ \textbf{to} $K-1$
       \textbf{step} $\mathit{blockDim.x}$}
    \If{$\mathit{topk\_id}[k]\neq\tau_{\mathrm{next}}$}
      \State \textbf{continue}
    \EndIf
    \State $\rho\leftarrow\mathit{topk\_logp}[k]$
    \If{$\rho\leq\theta_{\mathrm{tok}}$ \textbf{or}
        $s.\sigma+\rho\leq\theta_{\mathrm{sent}}$}
      \State \textbf{break}
    \EndIf
    \State $s'\leftarrow\Call{BeamState}{\mathit{parent}{=}s,\,\mathit{token}{=}\tau_{\mathrm{next}}}$
    \State $s'.\sigma\leftarrow s.\sigma+\rho$
    \State $s'.\mathit{link\_offset}\leftarrow s.\mathit{link\_offset}+1$;\;
           $s'.\mathit{end\_link}\leftarrow\mathit{end}$
    \If{$\mathit{end}$}
      \If{$\operatorname{is\_terminal}(s'.v)$}
        \State $\mathit{selected}.\operatorname{push}(s')$
      \EndIf
    \Else
      \State $\mathit{next}.\operatorname{push}(s')$
    \EndIf
    \State \textbf{break}
  \EndFor
\end{algorithmic}
\end{algorithm}

\begin{algorithm}
\caption{ExtendNode (device; one CTA per parent beam)}
\label{alg:extendnode}
\begin{algorithmic}[1]
  \State Compute child range $[\mathit{base\_node},\,\mathit{base\_node}+\mathit{deg})$
         via LOUDS $\operatorname{select}_0$ and \Call{WarpNextUnset}{}
  \If{$\mathit{deg}=0$}
    \Return
  \EndIf
  \For{$k=\mathit{threadIdx.x}$ \textbf{to} $K-1$
       \textbf{step} $\mathit{blockDim.x}$}
    \State $\tau\leftarrow\mathit{topk\_id}[k]$;\;
           $\rho\leftarrow\mathit{topk\_logp}[k]$
    \If{$\rho\leq\theta_{\mathrm{tok}}$ \textbf{or}
        $s.\sigma+\rho\leq\theta_{\mathrm{sent}}$}
      \State \textbf{continue}
    \EndIf
    \State Binary-search sorted child labels in
           $[\mathit{base\_node},\,\mathit{base\_node}+\mathit{deg})$ for $\tau$;
           peek into $\mathit{tail}$ when the visited child is a link node
    \If{not found}
      \State \textbf{continue}
    \EndIf
    \State $s'\leftarrow\Call{BeamState}{\mathit{parent}{=}s,\,\mathit{token}{=}\tau}$;\;
           $s'.\sigma\leftarrow s.\sigma+\rho$;\;
           $s'.v\leftarrow$ matched child
    \State Route $s'$ to $\mathit{selected}$ or $\mathit{next}$ per
           terminal- and link-status (same rules as Algorithm~\ref{alg:extendinlink})
  \EndFor
\end{algorithmic}
\end{algorithm}

\subsection{Warp-Level LOUDS Navigation}
\label{subsec:warp_louds}

The inner step of child enumeration, finding the next $0$-bit in
\texttt{louds\_} starting from a given position, can span many
64-bit words when a node has high out-degree, and a sequential
per-thread scan leaves 31 of the 32 lanes of a warp idle. FlashTrie
uses a {warp-ballot} loop in which each lane checks one of the
next 32 words simultaneously and the warp votes via
\texttt{\_\_ballot\_sync}; the position of the first word containing a
$0$-bit is recovered by \texttt{ffs} on the ballot mask, and the
final intra-word bit position by another \texttt{ffs} on the negated
word. Asymptotically this turns the $O(\Delta)$ inner loop into
$O(\Delta/32)$ memory accesses. Algorithm~\ref{alg:warpnextunset}
gives the exact word-level masking and ballot pattern; the lane
arithmetic is what prose cannot specify precisely.

\begin{algorithm}
\caption{WarpNextUnset (device; 32-thread warp)}
\label{alg:warpnextunset}
\begin{algorithmic}[1]
  \State $w\leftarrow\lfloor i/64\rfloor$;\;
         $\mathit{word}\leftarrow B[w]\mid{\sim}({\sim}0\ll(i\bmod 64))$
    \Comment{mask bits below $i$}
  \If{$({\sim}\mathit{word})\neq 0$}
    \Return $w\cdot 64+\operatorname{ffs}({\sim}\mathit{word})-1$
  \EndIf
  \State $w\leftarrow w+1$;\;
         $\mathit{lane}\leftarrow\mathit{threadIdx.x}\bmod 32$
  \Repeat
    \State $\mathit{ballot}\leftarrow\operatorname{ballot}(({\sim}B[w+\mathit{lane}])\neq 0)$
    \If{$\mathit{ballot}=0$}
      \State $w\leftarrow w+32$
    \Else
      \State $w\leftarrow w+\operatorname{ffs}(\mathit{ballot})-1$;\;\textbf{break}
    \EndIf
  \Until{false}
  \Return $w\cdot 64+\operatorname{ffs}({\sim}B[w])-1$
\end{algorithmic}
\end{algorithm}

\subsection{Top-$K$ Selection on the GPU}
\label{subsec:gpu_topk}

After expansion the candidate buffer \texttt{next} is trimmed to
the top-$B$ hypotheses by length-normalized score. FlashTrie
switches between two implementations depending on $BW$.

\paragraph{Block merge sort ($B \leq 1024$).}
For small $BW$ the entire selection fits in a small number of
sort-blocks of size $2B$. Each CTA sorts one block in descending
order of $\hat\sigma$ using CUB's \texttt{BlockMergeSort}, then the
sorted blocks are combined by a tournament-merge tree in
$\lceil \log_2 (n / 2B) \rceil$ rounds with a grid-wide barrier
between rounds. After the final merge round the first $BW$ entries
are the top-$B$ beams. Total work is $O(n \log n / P)$ for the
initial sort and $O(B \log(n/B) / P)$ for the merge rounds, with
$P$ the grid-wide thread count; see
Algorithm~\ref{alg:blockmergesort}.

\begin{algorithm}
\caption{BlockMergeSort Top-$K$ (device; all blocks)}
\label{alg:blockmergesort}
\begin{algorithmic}[1]
  \State $\mathit{remaining}\leftarrow\lceil|\mathit{cand}|/(2B)\rceil$
    \Comment{number of sort-blocks}
  \While{$\mathit{remaining}\geq 1$}
    \Comment{Phase 1: sort each block of $2B$ elements}
    \ForAll{block $b$ assigned to this CTA}
      \State Load $2B/P$ elements per thread (pad with $-\infty$ if OOB)
      \State $\Call{BlockMergeSort}{\mathit{thread\_data}}$ (descending by $\hat\sigma$)
      \State write back to $\mathit{sort\_state}$
    \EndFor
    \If{$\mathit{remaining}=1$}
      \State \textbf{break}
    \EndIf
    \State $\operatorname{sync}(\mathcal{G})$
    \Comment{Phase 2: tournament merge}
    \State $\mathit{remaining}\leftarrow\lceil(\mathit{remaining}-1)/2\rceil+1$
    \State $\operatorname{sync}(\mathcal{G})$
  \EndWhile
  \Comment{Phase 3: collect top-$B$ entries into $\mathit{cand}[0\ldots B-1]$}
\end{algorithmic}
\end{algorithm}

\paragraph{Output serialization.}
The final \texttt{selected} buffer is descending-sorted by
$\hat\sigma$ but the wire format expects ascending order, so
\textsc{GenerateResult} writes the $i$-th selected beam into slot
$n_o {-} 1 {-} i$ of the output arrays. Per-beam token sequences
are reconstructed by walking the \texttt{beam\_history} parent
pointers in reverse, with a CUB \texttt{BlockScan} converting
per-beam lengths into the end-offsets that index the flat
\texttt{tokens} buffer. The on-wire layout for $n_o$ beams totalling
$n_{\mathrm{tok}}$ tokens is an 8-byte \texttt{num\_beams} header
followed by three contiguous arrays: $8 n_o$ bytes of
\texttt{logp\_norm} (\texttt{float64}), $8 n_o$ bytes of
\texttt{sent\_offset} (\texttt{uint64}), and $4 n_{\mathrm{tok}}$
bytes of \texttt{tokens} (\texttt{uint32}); sequence $i$ occupies
\texttt{tokens[sent\_offset[$i{-}1$] : sent\_offset[$i$]]}.

\paragraph{Per-batch GPU memory budget.}
A single search slot occupies roughly $320$\,MB of GPU memory,
dominated by the persistent device-resident state: three frontier
arrays at $56$\,MB each, a $56$\,MB beam history, six $8$\,MB
sort-state arrays, a $\sim$$128$\,MB device-side output buffer
paired with pinned host memory, and a $\sim$$12$\,MB input buffer
similarly paired. Top-$K$ scratch is small ($3 \times 0.5$\,MB) and
the trie struct shell is a few hundred bytes; the trie arrays
themselves live in Unified Memory and range from 50 to 500\,MB
depending on the constraint set, charged once across all slots
rather than per-slot. A four-slot batch therefore uses
$\approx 1.3$\,GB plus the trie, which fits comfortably alongside a
generation-model weights footprint on a single 80\,GB device.

\subsection{Runtime Breakdown Methodology}
 \label{app:runtime_breakdown}
 
 Because beam search runs as a single persistent cooperative kernel
 (Section~\ref{subsec:fm}), we measure its internal phase breakdown
 with lightweight on-device instrumentation rather than separate
 kernel launches. Under a compile-time flag, a single grid-leader
 thread reads the GPU cycle counter (\texttt{clock64}) along its
 timeline. We time the \emph{wall-clock span} of each phase region,
 bracketed by the grid barriers (\texttt{grid.sync}) that already
 separate the phases; because each region ends at a barrier, its span
 includes the slowest CTA, i.e.\ the phase's critical-path time across
 the grid rather than the leader's own work alone. Cycle counts are
 converted to milliseconds via the SM clock rate and averaged over the
 13{,}000-request workload.
 
 The four buckets map to kernel regions as follows. \emph{Expansion}
 is the per-proposal binary search that locates a candidate token
 among a parent's children, including the dependent LOUDS/tail memory
 accesses at each search step (a single tail read for link-interior
 beams). \emph{Validation} is per-candidate admission: the per-token
 and cumulative log-probability threshold tests, the terminal-node
 check, and the append into the candidate buffer. \emph{Pruning} is
 the top-$B$ selection (parallel mergesort) that
 enforces the beam width. \emph{Sync} is the residual: inter-phase
 grid barriers outside the work regions, frontier setup/save, result
 serialization, and buffer swaps. The expansion and validation regions
 are interleaved per proposal, so we report their combined wall span
 split by the leader thread's measured compute ratio; pruning and sync
 are measured directly. The four phases sum to total kernel time by
 construction.
 
 We report \emph{proportions} rather than absolute times: the
 instrumented build incurs mild register pressure that inflates
 absolute latency, so all latency numbers in
 Section~\ref{subsec:latency_cpu} come from a separate,
 non-instrumented build. For reference, the measured mean per-phase
 times at $b{=}1$ range from $0.34$/$0.07$/$0.08$/$0.03$\,ms
 (expansion/validation/pruning/sync) at $BW{=}100$ to
 $0.66$/$0.79$/$0.35$/$0.07$\,ms at $BW{=}1000$, with total kernel time
 under $2$\,ms across the full sweep.
 
\section{Experimental Setup: Full Protocol}
\label{app:setup_detail}

This appendix fills in the measurement details abbreviated in the
main body's Experimental Setup
(Section~\ref{sec:methodology}/Section~\ref{subsec:setup_results}). The
deployment envelope (1$\times$ A100 80\,GB, AMD EPYC 7V13 host,
60\,GB host RAM, NUMA-pinned to cores 0--7 of socket~0 via
\texttt{numactl --cpunodebind=0 --membind=0 taskset -c 0-7},
Docker container) and the latency-window definition (C++
entry-point timer covering H2D, kernel, and D2H with stream
synchronisation) are reused from the main body without
modification.

\paragraph{Comparability of the timing window.}
The CPU systems' timer closes when the search call returns; the
FlashTrie timer closes after \texttt{cudaStreamSynchronize}.
All three systems share the same C++ entry-point boundary and
exclude only language-binding marshalling, so reported latencies
are directly comparable. NUMA-pinning to a single socket also
removes cross-socket traffic on the H2D path, so FlashTrie's
speedup is reported under the same deployment-realistic envelope
as the CPU baselines.

\paragraph{Length-normalised scoring.}
Beams are ranked by the Google-style length-normalised
log-probability used in
Section~\ref{subsec:setup}~\cite{wu2016googles}:
\begin{equation}
s_{\text{norm}}(\mathbf{x}_{0:d}) =
\Bigl(\sum_{j=0}^{d-1} \ell_{j,\sigma(j)}\Bigr)
\cdot \Bigl(\frac{6}{5+d}\Bigr)^{\alpha},
\label{eq:length_norm}
\end{equation}
where $\alpha$ controls the normalisation strength and $\sigma(j)$
indexes the selected candidate at step $j$.

\paragraph{Per-request latency sampling.}
Each $(BW, b)$ configuration is benchmarked over the full
13{,}000-request dataset at batch size~1. One warm-up pass is
discarded; FlashTrie additionally runs a short GPU warm-up
before the measurement window opens. The next 10 passes are
timed and pooled, producing $130{,}000$ samples per configuration
from which the mean and the 50th/90th/95th/99th percentiles are
computed. The across-run standard deviation is reported as a
variance bar in figures and as the $\pm$ value in
Table~\ref{tab:latency_percentiles}.

\paragraph{Throughput sweep.}
For the throughput evaluation
(Figure~\ref{fig:throughput}, Section~\ref{sec:results}) we sweep
$b \in \{1, 4, 8\}$ at
$BW \in \{100, 300, 500, 700, 1000\}$. For each $(K,b)$ the dataset
is run as $\lceil 13000 / b \rceil$ batched calls per pass, and
per-pass throughput is total queries divided by total wall-clock
duration measured with the same C++ entry-point timer window used
for latency. One warm-up pass is discarded and four passes are
timed; reported $\Theta$ is the mean over those four passes, with
the across-pass standard deviation drawn as the error band in
Figure~\ref{fig:throughput}.

\paragraph{Precision@K scoring.}
For the retrieval-quality evaluation (Section~\ref{subsec:precision})
each request's top-$K$ trie-admitted output is scored by a
transformer teacher model adapted from ~\cite{valluri2025pixnar} that takes the request and a candidate
keyword sequence as input and returns a single relevance score.
The teacher is identical across all three systems
(MARISA-Opt, FlashTrie, and PPT) so that any precision
differences are attributable to the candidate set, not to the
scorer. Precision@$BW{=}100$ and Precision@$BW{=}200$ are reported
as the macro-average over the 13{,}000 requests.


\section{Per-Percentile Latency Table}
\label{app:latency_percentiles}

Table~\ref{tab:latency_percentiles} gives the full per-percentile
breakdown of the per-request trie-search latency reported in
Figure~\ref{fig:latency_cpu_gpu}: mean, p50, p90, p95 and p99 in
milliseconds, for every beam width $BW$, computed as the median
across 10 runs of 13{,}000 queries each. 

\begin{table*}[tbp]
  \centering
  \setlength{\abovecaptionskip}{2pt}
  \setlength{\belowcaptionskip}{2pt}
  \caption{Per-request trie-search latency (ms), full percentile
    breakdown. Each cell is the median across 10 runs of 13{,}000
    queries each. \textbf{Lower is better.}}
  \label{tab:latency_percentiles}
  \footnotesize
  \setlength{\tabcolsep}{6pt}
  \renewcommand{\arraystretch}{0.9}
  \begin{tabular}{@{}c rrrrr rrrrr@{}}
     \toprule
    & \multicolumn{5}{c}{\textbf{MARISA-Opt (CPU)}}
    & \multicolumn{5}{c}{\textbf{FlashTrie (GPU)}} \\
    \cmidrule(lr){2-6}\cmidrule(lr){7-11}
    \textbf{$BW$}
       & Mean & p50 & p90 & p95 & p99
       & Mean & p50 & p90 & p95 & p99 \\
    \midrule
     100 &  9.0 &  8.9 & 11.8 & 13.0 & 15.1
         &  0.6 &  0.5 &  0.6 &  0.6 &  0.7 \\
     200 & 14.9 & 14.4 & 19.2 & 20.9 & 23.6
         &  0.7 &  0.7 &  0.8 &  0.9 &  0.9 \\
     300 & 20.1 & 19.8 & 25.2 & 26.9 & 30.0
         &  0.9 &  0.9 &  1.1 &  1.1 &  1.2 \\
     400 & 25.3 & 25.1 & 32.2 & 34.2 & 38.3
         &  1.1 &  1.1 &  1.4 &  1.5 &  1.6 \\
     500 & 29.5 & 29.3 & 38.2 & 40.6 & 45.4
         &  1.4 &  1.4 &  1.7 &  1.8 &  2.0 \\
     600 & 33.4 & 33.0 & 44.2 & 47.4 & 53.9
         &  1.6 &  1.5 &  2.0 &  2.2 &  2.6 \\
     700 & 36.3 & 35.9 & 48.8 & 52.6 & 60.4
         &  1.6 &  1.6 &  2.2 &  2.4 &  2.8 \\
     800 & 40.5 & 38.4 & 52.7 & 57.0 & 66.5
         &  1.9 &  1.8 &  2.5 &  2.7 &  3.2 \\
     900 & 42.9 & 40.6 & 55.9 & 60.5 & 70.5
         &  1.9 &  1.9 &  2.5 &  2.8 &  3.3 \\
    1000 & 46.3 & 43.8 & 60.6 & 65.5 & 76.7
         &  1.9 &  1.8 &  2.5 &  2.8 &  3.3 \\
    \bottomrule
  \end{tabular}
\end{table*}


\section{Online A/B Test Protocol}
\label{app:online_flight_protocol}

This appendix gives the  operational protocol of the
production A/B test summarised in
Section~\ref{subsec:online_flight}.

\paragraph{Deployment motivation.}
Offline benchmarks
(Sections~\ref{subsec:latency_cpu}--\ref{subsec:precision})
establish that FlashTrie preserves retrieval quality at a
fraction of the latency budget of CPU-resident MARISA-Opt. The
operative question for a monetised production system is whether
these system-level wins survive the realities of live traffic:
bursty query mixes, stale caches, downstream auction dynamics,
and revenue-sensitive guardrails. The flight is designed to
answer that question end-to-end.

\paragraph{Sample-ratio-mismatch (SRM) check.}
A sample-ratio-mismatch check passed throughout the flight,
confirming the realised treatment\,:\,control event ratio is
consistent with the configured allocation after eligibility
filtering. This rules out exposure-bias confounds before reading
the metric lifts.

\paragraph{Latency-timing boundary.}
End-to-end latency is measured from request ingress at the
retrieval service to emission of the final top-$k$ candidate
set, inclusive of trie traversal, beam expansion, scoring, and
post-processing. Control and treatment share the entire
downstream stack (auction, pacing, allocation, click models);
they differ only in the retrieval-stage trie back-end and the
beam configuration the back-end can sustain within the $30$\,ms
end-to-end latency SLA.

\paragraph{Constraint library and back-end configurations.}
The production constraint library indexes hundreds of millions of
docIDs. The control arm runs
production CPU MARISA-Opt at $B_{\text{ctrl}}{=}200$,
$\text{top-}k{=}300$, with a 3-thread pool matched to the
per-worker MIG-slice CPU-core allocation.

\paragraph{Metric glossary.}
Headers in Table~\ref{tab:online_flight}: \emph{Revenue} is
total advertiser revenue for the arm;
\emph{Coverage} is the fraction of queries for which a
sponsored ad is shown; \emph{Impression} is Impression Yield
(IY), the expected value per impression opportunity; \emph{DefectML-FBS}
aggregates the Defect-ML classifier and Feedback-Based-Suppression
(FBS) guardrails on the Ads Selection.

\section{MARISA-Int Feasibility Measurements}
\label{app:marisa_int}

MARISA-Int is included only as a feasibility check (not a primary performance
baseline). At billion-key scale it is too slow for a full 13K sweep, so we
report a fixed 10-query subsample (seed 42) and omit $K\ge500$ after observing
a single $BW{=}1000$ query at $10{,}997$\,s ($\approx$3.05\,h).

Table~\ref{tab:marisa_int} reports mean/p50/p90/p95 on the subsample. p99 is
omitted at $N{=}10$. The large mean--median gap reflects high per-query
variation in traversal depth and root branching.

\begin{table}[tbp]
  \centering
  \caption{%
    MARISA-Int per-request latency on a fixed 10-query random
    subsample of the 13{,}000-request workload (seed 42, identical
    query indices across $BW$).
  }
  \label{tab:marisa_int}
  \setlength{\tabcolsep}{4pt}

  \begin{tabular}{c rrrr}
     \toprule
      & \multicolumn{4}{c}{\textbf{Latency (s)}} \\
      \cmidrule(lr){2-5}
      \textbf{$BW$}
        & \textbf{Mean} & \textbf{p50} & \textbf{p90} & \textbf{p95} \\
      \midrule
        100  &     394.5 &     37.6 & 1{,}049.7 & 1{,}809.4 \\
        200  &     764.6 &    492.0 & 2{,}392.3 & 2{,}597.4 \\
        300  &  1{,}498.9 &    947.7 & 4{,}333.6 & 5{,}134.4 \\
        400  &  2{,}028.9 & 1{,}329.3 & 4{,}789.9 & 6{,}391.6 \\
      \bottomrule
  \end{tabular}
\end{table}

Latency grows roughly linearly with $BW$, consistent with MARISA-Int's serial
per-beam execution model, and confirms it is unsuitable as a main comparator at
this operating point (hence MARISA-Opt in Section~\ref{subsec:iom}).

\section{PPT Baseline Construction}
\label{app:ppt_details}

For each decoding depth $d \in [0, T)$, we collect the set of all token
IDs that appear at position~$d$ in any keyword of the constraint library
and store them as a sorted, deduplicated array on the GPU. A flat
\texttt{(valid\_tokens, valid\_offsets, valid\_counts)} layout is used so
that the depth-$d$ alphabet is contiguous in device memory. At search
time, for each beam at depth $d$, every top-$K$ proposal token is tested
against $\texttt{valid\_tokens}[d]$ via a single \texttt{lower\_bound}
binary search; tokens that pass are admitted into the next beam, tokens
that fail are pruned. The cooperative kernel, beam scoring, log-prob
thresholds, length normalisation, and top-$B$ pruning are unchanged from
FlashTrie, the only difference is the constraint structure.

PPT admits a token $x_{t,i}$ at depth $d$ iff $x_{t,i}$ appears at
position~$d$ in some keyword; it does not require that the prefix
$x_{0:d}$ extended by $x_{t,i}$ corresponds to a valid path in the
constraint library. The candidate set produced by PPT is therefore a
strict superset of the trie-valid set, and PPT performs strictly less
work per step than FlashTrie.


\section{Full Ablations: Binary-Search and Linear-Search Variants}
\label{app:ablations_full}

This appendix gives the full per-$BW$ data for the two ablations
summarised in Table~\ref{tab:ablations} (PPT and the per-key linear
probe), and then reports two additional, fully sequential GPU
variants that were measured but failed to scale to the production
benchmark. Together with the main-text PPT ablation
(Section~\ref{subsec:ablations}), these results decompose the
FlashTrie speedup into three additive contributions: GPU
residency and kernel fusion (PPT vs MARISA-Opt), the LOUDS-trie
structure for path validity (FlashTrie vs PPT), and binary
search at the inner per-key loop (FlashTrie vs the per-key
linear probe).

\subsection*{Per-$BW$ Sweep: PPT and Per-Key Linear Probe}

Table~\ref{tab:ablations_full} gives the complete per-$BW$ sweep
for the two ablations summarised in Table~\ref{tab:ablations}:
all ten beam widths $BW \in \{100, 200, \dots, 1000\}$ on the same
13K-request workload, with Mean and p95 latency in milliseconds
and the speedup column equal to ablation\,p95 divided by
FlashTrie\,p95 at the same $BW$.

\begin{table}[tbp]
\centering
\caption{Full per-$BW$ ablation sweep (companion to
  Table~\ref{tab:ablations}). \textbf{Lower is better.}}
\label{tab:ablations_full}
\scriptsize
\setlength{\tabcolsep}{2pt}
\resizebox{\columnwidth}{!}{%
\begin{tabular}{c rr c rr c}
   \toprule
  & \multicolumn{3}{c}{\textbf{PPT}}
  & \multicolumn{3}{c}{\textbf{Linear-probe}} \\
  \cmidrule(lr){2-4} \cmidrule(lr){5-7}
    \textbf{$BW$}
       & Mean & p95 & \textbf{Speedup}
       & Mean & p95 & \textbf{Speedup} \\
     \midrule
      100  & 0.57  & 0.79  & $1.23\times$  & 110.2 & 134.1 & $209\times$ \\
      200  & 1.16  & 1.61  & $1.85\times$  & 119.6 & 144.2 & $166\times$ \\
      300  & 1.90  & 2.69  & $2.40\times$  & 123.9 & 149.1 & $133\times$ \\
      400  & 2.73  & 3.97  & $2.71\times$  & 126.8 & 152.8 & $104\times$ \\
      500  & 3.54  & 5.20  & $2.84\times$  & 128.6 & 155.7 & $85\times$  \\
      600  & 5.57  & 8.22  & $3.69\times$  & 135.5 & 169.3 & $76\times$  \\
      700  & 6.98  & 10.30 & $4.37\times$  & 141.1 & 178.8 & $76\times$  \\
      800  & 8.22  & 12.15 & $4.45\times$  & 145.6 & 186.3 & $68\times$  \\
      900  & 9.73  & 14.36 & $5.21\times$  & 151.2 & 194.3 & $71\times$  \\
     1000  & 11.28 & 16.65 & $5.98\times$  & 154.3 & 198.7 & $71\times$  \\
     \bottomrule
\end{tabular}%
}
\end{table}

\subsection*{Why the Inner Search Algorithm Matters: Linear-Search Variants}
\label{app:linear_ablation}

The PPT ablation attributes the remaining gap between FlashTrie
and a GPU baseline to the LOUDS-trie data structure. Within the
trie kernel itself, however, the per-parent child lookup is
performed by a binary search over the parent's sorted child labels.
To isolate the contribution of this algorithmic choice, as distinct
from GPU residency, kernel fusion, and the LOUDS layout, we
replaced the binary-search inner loop with three successively
weaker variants and measured each in turn. We report this ablation
primarily as evidence that the seemingly small step from binary to
linear search at the inner loop is responsible for a substantial
fraction of FlashTrie's wall-clock advantage, and as a
cautionary data point for naive GPU ports of the textbook MARISA
traversal.

All three variants share FlashTrie's cooperative kernel, beam
scoring, log-prob thresholds, and length normalisation. They differ
only in how each parent beam locates the children whose labels
appear in the top-$K$ proposal set:
\begin{itemize}
  \item \textbf{Two-pointer merge (vanilla MARISA port).} Faithful
    single-thread port of the CPU MARISA child-extension loop:
    one CTA per parent, one thread per CTA walks the parent's
    sorted children and the sorted top-$K$ in lockstep,
    $O(\text{degree} + K)$ per parent. The remaining 511 threads
    of the CTA idle. Across-parent parallelism is preserved
    (one CTA per parent, many CTAs per SM), matching the CPU
    pattern of multi-threading across parents.
  \item \textbf{Per-child parallel linear scan.} Block-parallel
    baseline that distributes children across the CTA's threads:
    thread $t$ handles children $t,\ t+\text{blockDim},\ldots$
    and for each assigned child performs an $O(K)$ linear scan
    through the top-$K$ array to test membership. Per-parent
    work $O(\text{degree} \cdot K / \text{blockDim})$ in
    wall-clock, but every thread re-issues trie metadata loads
    (\texttt{select0}, \texttt{rank1}, label decode) for its
    assigned children with no sharing.
  \item \textbf{Per-key linear probe (FlashTrie inner loop with
    binary to linear).} Identical parallelisation to
    FlashTrie -- one thread per top-$K$ key, striding
    \texttt{blockDim} -- but each thread linearly scans the
    parent's sorted children for its assigned key (with
    sorted-array early exit) instead of binary searching. To make
    this competitive at all, the CTA cooperatively decodes each
    parent's child labels once into a shared-memory cache, after
    which all 512 threads scan the cache. Per-parent work
    $O(\text{degree} + K \cdot \text{degree} / \text{blockDim})$
    in arithmetic, with all global-memory trie loads amortised
    across the top-K loop.
\end{itemize}

Both fully sequential variants fail to scale to the full benchmark
at production beam widths. The two-pointer merge, although
algorithmically the lowest in arithmetic complexity per parent
($O(\text{degree}+K)$), uses only $1/512$ of each CTA's threads;
the resulting underutilisation makes the warm-up pass alone (20
iterations on a single batch) take several minutes at $K=100$, with
a full 13K-query pass projected at well over an order of magnitude
longer than even the per-child parallel variant. The per-child
parallel scan is similarly impractical: although every thread is
busy, the redundant per-thread re-decoding of children labels
(each thread re-issues \texttt{select0}, \texttt{rank1}, and label
lookups for its slice) and the $O(K)$ inner scan per child push
per-batch latency into the multi-second range. A single 13K-query
pass at $BW=100$ failed to complete within seven hours, ruling out
the variant for our sweep.

Only the per-key linear probe with cooperative shared-memory
caching admits a meaningful comparison: it shares FlashTrie's
parallelisation scheme exactly, differing only in whether the inner
per-key search over the cached children is binary or linear. Even
so, at $BW{=}100$ the per-key linear probe has a mean per-request
latency of 110.2~ms against FlashTrie's 0.55~ms
(${\sim}200\times$ slower; full per-$BW$ sweep in
Table~\ref{tab:ablations_full}), indicating that binary search at
the inner loop contributes a two-orders-of-magnitude factor on top
of the GPU residency, kernel fusion, and shared-memory caching
that all three variants share.

The first two variants illustrate a common failure mode of porting
CPU succinct-trie code to the GPU. The CPU MARISA loop is fast
because (i) modern x86 cores are individually fast and (ii)
MARISA-Opt multi-threads across parents to exploit thread-level
parallelism. Both properties evaporate on the GPU: an SM's
individual thread is far weaker than a CPU core, and the
across-parents axis alone is not wide enough to amortise the trie
metadata cost when each parent's inner loop is sequential. The
per-child parallel variant uses all threads but pays for it with
redundant global-memory traffic that swamps the arithmetic
savings. Only by holding the parallelisation scheme fixed and
varying the inner-loop algorithm, the per-key linear probe
versus FlashTrie's binary search, can the contribution of
the search algorithm itself be measured cleanly, and that
contribution is large.


\section{NQ + GENRE Workload Construction and Realistic-Threshold Analysis}
\label{app:nq_genre_setup}
\label{app:realistic_thresholds}

The three components introduced in
Section~\ref{subsec:public_data} are pinned for exact reproduction:
NQ-Open validation (\textbf{3{,}600 questions}, free-form input
matching serving shape), the public GENRE-KILT checkpoint
(\texttt{facebook/genre-kilt}, BART-large with a ${\sim}50$k token
vocabulary, ungated), and the deduplicated KILT Wikipedia titles
(\textbf{${\sim}6$M tokenized titles}, BART-tokenized so the model's
emitted token IDs and the trie's edge labels share one ID space,
avoiding train/serve vocabulary mismatch).

GENRE is autoregressive: a vanilla top-$K$ beam roll-out produces a
proposal grid where each position's top-$K$ is conditioned on the
previous-step argmax, so the grid is internally coherent and the trie
does little admissibility filtering. To simulate the cross-position
incoherence of production NAR grids, we use a two-stage pipeline.

\emph{Stage 1: AR top-$K$ extraction.}
For each NQ-Open query we run GENRE over $T{=}16$ decoding positions,
emitting at each position $t$ the top-$K{=}600$ tokens with their
log-softmax scores. The result is a per-query $(T, K)$ grid of token
IDs and log-probs, the same data shape FlashTrie and MARISA-Opt
consume on internal traffic.

\emph{Stage 2: head + noise-tail replacement.}
For a chosen head size $r \in \{0, 9, 18, 37, 75, 150, 300, 600\}$
we keep the top-$r$ AR-emitted tokens at each position (the
``coherent head'') and replace ranks $[r, K)$ with tokens sampled
from the position-stratified marginal distribution of the trie corpus
(the empirical frequency of each token at position $t$ across all
tokenized Wikipedia titles). Noise tokens inherit the AR model's tail
log-probs at the occupied ranks so the resulting grid is
indistinguishable from a real model emission to a downstream beam
scorer. The knob $r$ interpolates between $r{=}0$ (pure NAR-style
proxy, maximum trie work) and $r{=}K$ (pure AR, minimum trie work).

Alternative noise designs (uniform-vocab, uniform-position alphabet, and
history-conditioned sampling) were rejected because they either under-stress the
trie or reintroduce AR coherence. Position-stratified marginal sampling
preserves per-position plausibility while remaining cross-position independent.
We keep AR tail log-probs to preserve realistic score decay under
$\theta_{\mathrm{tok}}$, $\theta_{\mathrm{sent}}$, and length normalization.

All experiments use the same Docker container as
Section~\ref{subsec:setup_results} (1$\times$ A100 80\,GB, 8 CPU cores
pinned via \texttt{numactl --cpunodebind=0 --membind=0 taskset -c
0-7}, 60\,GB host RAM). The constraint trie is built once over the
tokenized KILT corpus and reused across all $r$. For the $r$-sweep,
beam-score thresholds are set permissively
($\theta_{\mathrm{tok}}{=}{-}20$, $\theta_{\mathrm{sent}}{=}{-}200$,
length-norm exponent $5.0$) so that the trie's admissibility check,
not the scorer, drives all pruning. Results are reported at
$BW{=}600$, which sits inside the beam-width range of the internal
sweep (Figure~\ref{fig:latency_cpu_gpu}) and has the densest $r$-grid
in the released artifacts. The latency measurement protocol
(Section~\ref{subsec:setup_results}) is reused verbatim.

\begin{table}[tbp]
  \centering
  \caption{Mean trie-search latency (ms) at $BW{=}600$ on NQ+GENRE
    vs.\ head size $r$ ($r{=}0$: NAR-style proxy, max trie work;
    $r{=}600$: GENRE verbatim). Speedup = CPU\,/\,GPU mean.
    \textbf{Lower latency, higher speedup are better.}}
  \label{tab:latency_vs_r}
  \small
  \setlength{\tabcolsep}{3.5pt}
  \begin{tabular}{l|*{6}{c}}
     \toprule
    \textbf{$r$}                & 0    & 9    & 18   & 150  & 300  & 600  \\
    \midrule
    \textbf{MARISA-Opt (ms)}    & 41.2 & 40.5 & 40.2 & 37.6 & 33.7 & 20.4 \\
    \textbf{FlashTrie (ms)}  &  5.7 &  5.2 &  3.6 &  2.8 &  2.7 &  2.4 \\
    \textbf{Speedup (CPU/GPU)} &  7.3 &  7.8 & 11.1 & 13.3 & 12.5 &  8.4 \\
    \bottomrule
  \end{tabular}
\end{table}

\subsection*{GPU Overhead Floor}

The GPU curve flattening in Table~\ref{tab:latency_vs_r} is explained by a
structural overhead floor (kernel launch, grid barriers, fixed top-$B$
selection passes, and fixed-size H2D/D2H I/O). CPU has less fixed overhead, so
its latency keeps scaling with reduced trie work while GPU saturates once work
drops below this floor.

\subsection*{Realistic-Threshold Operating Point}

Table~\ref{tab:latency_vs_r} uses permissive thresholds to stress trie work.
Here we evaluate realistic production thresholds (Section~\ref{subsec:online_flight}),
where scorer-side pruning is stronger under the same 30\,ms SLA.

We keep all other NQ+GENRE settings fixed (3,600 queries, $BW{=}600$, $T{=}16$,
length norm 5.0, identical warmup) and use the deployed threshold pair.
Table~\ref{tab:realistic_thresholds} reports mean latency and CPU/GPU speedup
across the $r$ sweep.

At realistic thresholds and worst-case $r{=}0$, MARISA-Opt drops from
41.19\,ms (permissive) to 8.95\,ms mean and FlashTrie from 5.66\,ms to 1.54\,ms,
showing scorer-side pruning reduces absolute work for both systems. Relative
advantage remains multi-$\times$: 5.8$\times$ at $r{=}0$, typically
7.4--8.3$\times$ for $r\in\{9,37,150,300\}$ (Table~\ref{tab:realistic_thresholds}).

Across the $r$ sweep, CPU latency still tracks workload coherence, while GPU
latency stays in a narrow band because fixed cooperative-kernel overhead
dominates once useful trie work is small. This preserves FlashTrie's core value
at production thresholds: lower and more stable latency under workload variance.

\begin{table}
  \centering
  \caption{%
    Per-request mean trie-search latency (ms) at $K{=}600$ under
    realistic production thresholds on the NQ + GENRE public
    workload, across the head-size sweep
    $r \in \{0, 9, 18, 37, 75, 150, 300, 600\}$. MARISA-Opt is
    the CPU baseline (8 worker threads, NUMA-local pinning);
    FlashTrie is the GPU implementation on a single A100
    80\,GB. The \textbf{Speedup} column reports CPU mean /
    GPU mean per row. Compared to the permissive-threshold sweep
    in Table~\ref{tab:latency_vs_r}, score-driven pruning ahead
    of the trie reduces CPU mean latency by $\approx 5\times$
    and GPU mean by $\approx 2\text{--}4\times$, while the
    relative CPU-to-GPU mean speedup remains in the same
    $5$--$13\times$ band.
    \textbf{Lower is better for latency; higher is better for
    speedup.}}
  \label{tab:realistic_thresholds}
  \setlength{\tabcolsep}{6pt}
  \small
  \begin{tabular}{rccc}
     \toprule
    \textbf{$r$}
      & \textbf{MARISA-Opt}
      & \textbf{FlashTrie}
      & \textbf{Speedup} \\
    \midrule
      0 & $8.95$ & $1.54$ & $5.8\times$ \\
      9 & $8.61$ & $1.13$ & $7.6\times$ \\
     18 & $8.45$ & $2.83$ & $3.0\times$ \\
     37 & $8.37$ & $1.00$ & $8.3\times$ \\
    150 & $7.22$ & $0.95$ & $7.6\times$ \\
    300 & $6.78$ & $0.91$ & $7.4\times$ \\
    600 & $6.70$ & $1.40$ & $4.8\times$ \\
    \bottomrule
  \end{tabular}
\end{table}
\end{document}